\documentclass{article}
\usepackage{spconf,amsmath,graphicx}
\usepackage{amsfonts,bm}

\usepackage{amssymb,amsmath,bm}
\usepackage{multirow}
\usepackage{booktabs}
\usepackage{multicol}
\usepackage{array}
\usepackage{enumitem}
\usepackage{multirow, makecell}

\usepackage[disable]{todonotes} %

\newcommand{\myparagraph}[1]{\vspace{.4em} \noindent \textbf{#1}\ }

\global\hyphenpenalty=5000

\title{
Injecting Text in Self-Supervised Speech Pretraining
}
\name{Zhehuai Chen, Yu Zhang, Andrew Rosenberg, 
 Bhuvana Ramabhadran, Gary Wang, Pedro Moreno\thanks{Thanks to Yinghui Huang, Manasa Prasad, Jesse Emond and Ruoming Pang for many discussions and infratructure related assistance.}}
\address{Google, Inc.}
\begin{document}
\ninept
\maketitle
\begin{abstract}

Self-supervised pretraining for Automated Speech Recognition (ASR) has shown varied degrees of success. 
In this paper, we propose to jointly learn representations during pretraining from two different modalities: speech and text. The proposed method, {\it tts4pretrain} complements the power of contrastive learning in self-supervision with linguistic/lexical representations derived from synthesized speech, effectively learning from untranscribed speech and unspoken text. Lexical learning in the speech encoder is enforced through an additional sequence loss term that is coupled with contrastive loss during pretraining. We demonstrate that this novel pretraining method yields Word Error Rate (WER) reductions of 10\% relative on the well-benchmarked, Librispeech task over a state-of-the-art baseline pretrained with wav2vec2.0 only. The proposed method also serves as an effective strategy to compensate for the lack of transcribed speech, effectively matching the performance of 5000 hours of transcribed speech with just 100 hours of transcribed speech on the AMI meeting transcription task. Finally, we demonstrate WER reductions of up to 15\% on an in-house Voice Search task over traditional pretraining. Incorporating text into encoder pretraining is complimentary to rescoring with a larger or in-domain language model, resulting in additional 6\% relative reduction in WER.

\end{abstract}
\section{Introduction}
\label{sec:intro}

Self-supervised pretraining has been successful in several speech and language tasks. 
In ASR, these techniques have demonstrated the ability to effectively leverage large amounts of untranscribed speech (e.g.~\cite{zhang2020pushing}).
However, self-supervised pretraining needs to discover effective representations for speech recognition using only internally consistent representations.  While these representations can be learned with multiple views (objectives), there is no guarantee that the learned representation is optimal for any given task such as ASR, language identification or speaker verification tasks~\cite{baevski2020wav2vec,shen2020knowledge,xia2021self}. To wit, fine-tuning of the pretrained encoder for the given task is always necessary for optimal performance.

Unspoken text is complementary to un-transcribed speech in self-supervised learning. It is also much easier to collect than un-transcribed speech. Pretraining techniques such as MoCo~\cite{xia2021self}, Contrastive Predicting Coding (CPC)~\cite{wang2020contrastive,talnikar2021joint}, Autoregressive Perdictive Coding (APC)~\cite{chung2020generative}, SimCLR~\cite{jiang2020speech}, etc.,~ generalize using un-transcribed speech, however, they cannot leverage unspoken text, thereby limiting the power of the learned representations.

In this paper, we propose to jointly learn representations during pretraining from two different modalities, namely speech and text. We show that Text-to-Speech (TTS) can inject this lexical and phonetic information to the speech encoder during pretraining.  We propose {\it tts4pretrain}, a method to use synthesized speech during pretraining of the encoder. Central to this technique is the use of additional auxiliary decoder objectives such as phoneme, grapheme and word-piece sequence prediction.
These losses coupled with contrastive learning on real and synthesized speech help to inject lexical information in the speech encoder during pretraining.

The main contributions of this paper are:
\begin{itemize}
\item  A novel algorithm {\it tts4pretrain} to learn encoder representations from both un-transcribed speech and unspoken text, thus allowing for the explicit injection of lexical/phonetic/linguistic information in self-supervised pretraining.
\item A significant reduction in the amount of transcribed data needed for subsequent ``fine tuning'' to the domain or task at hand thereby directly resulting in cost savings.
\item A framework to adapt out-of-domain speech representations using in-domain text data through TTS.
\item Language-model fusion is complementary to the textual information introduced in pretraining. 
\item Generalization of the algorithm to different encoder architectures and sequence training objectives such as Connectionist Temporal Classification(CTC), Recurrent Neural Network Transducers(RNN-T), and Hybrd Autoregressve Transducers(HAT).

\end{itemize}

We present results on two publicly available, well-benchmarked ASR tasks, namely LibriSpeech and AMI meeting transcription tasks, and on queries representative of Google Voice Search traffic. We demonstrate that this novel pretraining method yields Word Error Rate (WER) reductions of 10\% relative on LibriSpeech over training with contrastive loss alone, establishing a new state of the art result. 
We also show that {\it tts4pretrain} matches the performance of 5,000 hours of transcribed speech with just 100 hours of transcribed speech on the AMI meeting transcription task. Finally, we demonstrate WER reductions of up to 15\% on an in-house Voice Search task over traditional pretraining.

The rest of this paper is organized as follows. 
We compare to related work in Section~\ref{sec:relate}. 
The proposed model is described in Section~\ref{sec:proposed}. Experiments are given in Section~\ref{sec:exp} with dataset and model details listed in Sections~\ref{sec:data} and \ref{sec:setup}. Ablation study is conducted in Section~\ref{sec:analysis}, followed by conclusion in Section~\ref{sec:conclude}.

\section{Related work}
\label{sec:relate}
Self-supervised pretraining techniques leverage untranscribed speech in ASR. wav2vec2.0~\cite{baevski2020wav2vec} has emerged as a successful training method that masks latent representations of input speech and solves a contrastive task over quantized speech representations. Recent advances in semi-supervised learning have revisited unsupervised learning in the form of Noisy Student Training (NST)~\cite{xie2020self,park2020improved} and introduced augmentation strategies such as FixMatch \cite{sohn2020fixmatch,weninger2020semi} and Sequential MixMatch~\cite{zhehuai2021} to ASR. Training methodologies to jointly learn from unpaired speech and text such as Deep Chain~\cite{tjandra2017listening}, cycle-consistency training~\cite{hori2019cycle}, and augmentation approaches~\cite{rosenberg2019speech, wang2020improving} are becoming increasingly popular. %

Leveraging vast amounts of unpaired text through learned text representations have been explored using shared encoder representations in~\cite{he2016dual, hayashi2018back, chen2020unpair}. These approaches have shown to be effective for ASR when combined with both transcribed and untranscribed speech~\cite{karita2019semi,ren2019almost}. Adversarial~\cite{liu2019adversarial} and cycle consistency training objectives~\cite{baskar2019semi} have also been proposed to leverage unpaired data.
Connectionist Temporal Classification (CTC) objective to train end-to-end models was first introduced in ~\cite{graves2006connectionist}. CTC has many advantages for ASR as it helps to improve robustness and achieve fast convergence~\cite{kim2017joint} and allows for streaming applications~\cite{moritz2019streaming}. Recurrent Neural Network Transducers (RNN-T)~\cite{graves2012sequence,graves2013speech} are also popular in streaming ASR applications. Both these objectives have been used in conjunction with unsupervised training.

Language model fusion in end-to-end ASR falls into two main approaches. These are approaches such as ``Shallow Fusion''~\cite{gulcehre2015using} that interpolate  scores from the end-to-end model and an external language model (LM) and approaches that jointly train end to end models and LMs, such as ``Cold Fusion''~\cite{sriram2017cold}, ``Deep Fusion''~\cite{gulcehre2015using}, ``Component Fusion''~\cite{shan2019component} and  
Hybrid Autoregressive Transducers (HAT)\cite{variani2020hybrid}. The HAT model separately preserves the internal LM learned by the E2E model thus allowing for a more accurate integration with an external LM. 
In this paper, we  propose a new method for combining untranscribed speech and synthesis of unspoken text in self-supervision (wav2vec2.0) with CTC and RNN-T training objectives. We also show that the proposed approach is complementary to both shallow fusion and HAT-based unspoken text integration.

\section{Proposed Method: tts4pretrain}
\label{sec:proposed}
\subsection{Framework and formulation}
\begin{figure}[hbt!]
  \centering
    \includegraphics[width=0.9\linewidth]{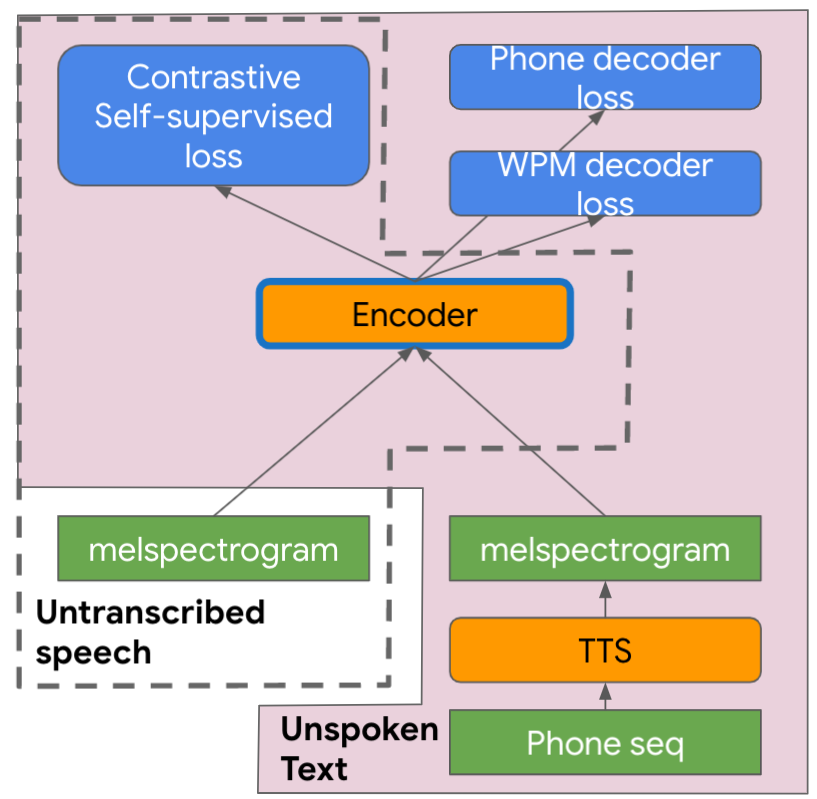}
    \caption{Proposed joint speech and text self-supervised pretraining architecture.}
    \label{fig:framework}
\end{figure}

\textit{Tts4pretrain} comprises two additional components that can be applied to any self-supervised pretraining techniques: 1) the use of synthesized utterances along with untranscribed ``real'' utterances during pretraining, and 2) the inclusion of auxiliary ASR-based losses.
Figure~\ref{fig:framework} shows the {\it tts4pretrain} framework.
In this paper, we follow the Wav2vec 2.0 pretraining framework  to apply contrastive loss on Conformer encoder representations. Every audio $x^*$ drawn from untranscribed speech corpora $\mathcal{L}_{speech}$ results in a loss function
$ \mathcal{J}_{\tt speech} = \mathcal{J}_{\tt w2v} (x^* \mid \theta_e) ,  x^* \in \mathcal{L}_{speech}$ used to optimize encoder parameters $\theta_e$.

To inject lexical information into the encoder, the pretraining data set includes synthetic utterances $x$ generated via speech synthesis (TTS) of text $y^*$ drawn from an unspoken text dataset $\mathcal{L}_{text}$.  The TTS model includes conditioning variables for both speaker conditioning, and a VAE-based latent variable for prosodic control (cf.~Section \ref{sec:tts}).  During synthesis these are sampled from $Z$, the set of appropriate conditional parameters (speaker embedding or VAE prior).
This results in a similar loss term for the synthesized utterances; 
\begin{eqnarray}
\label{eqn:text}
\begin{aligned}
\mathcal{J}_{\tt text} &=  \mathcal{J}_{\tt w2v} (x\mid \theta_e)\\
\ x &= \tt{TTS}(y^*,z), y^*\in \mathcal{L}_{text}, z \sim Z .
\end{aligned}
\end{eqnarray}

While self-supervision has been used to improve a variety of speech tasks,  our aim is to improve ASR performance.
Thus, we encourage the encoder to learn representations that useful for ASR by introducing supervision through auxiliary decoders $\theta_d$. The decoder objective function recognized the text $y^*$ of synthesized utterances $x =\tt{TTS}(\hat{x} \mid y^*,z)$.
In this case, both $\theta_e$ and $\theta_d$ are optimized, though only the encoder parameters $\theta_e$ are used in pretraining; $\theta_d$ is discarded.  We find that a linear readout layer followed by CTC loss is an effective auxiliary decoder, but any ASR decoder can be used here. (We evaluate other options in Section \ref{sec:decoders}.)  The auxiliary loss is defined as 
\begin{eqnarray}
\label{eqn:aux}
\begin{aligned}
\mathcal{J}_{\tt aux} &= \mathcal{J}_{\tt CTC} \left[ p(y^* | x, \theta_e,\theta_d)\right]\\
\ x &= \tt{TTS}(y^*,z), y^*\in \mathcal{L}_{text}, z \sim Z .
\end{aligned}
\end{eqnarray}
Note that the text labels, $y^*$ that are necessary for synthesis are available for auxiliary ASR loss calculation.  This is similar to the use of TTS in ASR training as in \cite{chen2020unpair,tjandra2018machine}, auxiliary ASR losses are also used in encoder training for voice conversion in \cite{biadsy2019parrotron}.

\subsection{On-the-fly Speech Synthesis and Utterance Selection}
\label{sec:tts}
We use a TTS system trained to generate ASR features from the unspoken text as the input of pretrained encoders.
TTS model is based on Tacotron 2D~\cite{jia2018transfer}, which takes text
sequences as input,
conditioned on speaker and utterance embeddings and outputs a sequence of mel spectrogram frames. 

Mel-filter bank features from the model can be consumed by the pretrained encoder,  eliminating the need for any vocoder.
To model prosody and increase its variability during inference, we use a hierarchical variational auto encoder (VAE) as in~\cite{hsu2018hierarchical,rosenberg2019speech}. This architecture captures local and global speaking styles separately and makes the  TTS more stable.
The hierarchical VAE includes a local encoder which encodes two-second chunks with a one-second overlap and a global encoder which encodes the whole utterance.

We follow \cite{chen2020unpair}, synthesizing distinct utterances on-the-fly during batch construction. Sampling a new $z \sim Z$ (speaker embedding and VAE latent) each time $y^*$ is included in a batch results in novel realizations of TTS utterances rather that training on the same TTS utterances during each training epoch.  

Previous work~\cite{hsu2021robust} analyzed the impact of the domain of untranscribed speech on self-supervised pretraining with the conclusion that in-domain untranscribed speech is more valuable that out-domain speech. %
To identify in- and near-domain text, we integrate the contrastive unspoken text selection method used in~\cite{chen2020unpair}.  This technique selects a subset of available sentences that is the most similar to a target domain.
The method requires two language models~\cite{moore2010intelligent,wang2019contrastive}: a background model $\mathbb{B}$, trained on any available unspoken text, and an in-domain model $\mathbb{D}$, trained on only in-domain material.  
We evaluate each sentence in the unspoken text corpus using the following equation:
\begin{equation}
\label{eqn:cds}
\mathcal S = \frac{\log P(\mathbf{w}| \mathbb{D}) - \log P(\mathbf{w}| \mathbb{B})}{\#(\mathbf{w})}
\end{equation}
where, the probabilities from the two language models are compared and normalized by the number of words to eliminate any length bias. We select the sentences with the top scores, $\mathcal S$, thus selecting sentences that are closer to the domain of $\mathbb{D}$.

\subsection{Contrastive Loss}
We pretrain a Conformer encoder \cite{gulati2020conformer} following \cite{zhang2020pushing}. We first use log-mel spectrograms from real data as input features and pass through 2 convolution subsampling blocks as a ``feature encoder'' to produce target frames (no quantization layer is used). The convolutional subsampling block has two 2D-convolution layers, both with strides $(2,2)$, resulting in a 4x reduction in the feature sequence length. A ``context network'' consists a stack of Conformer blocks makes predictions over the masked frames. A contrastive loss is optimized between the context vectors from the masked positions and the target context vectors. After the pretrained encoder converges on real untranscribed speech, we repeat the pretraining procedure on both TTS and real speech. Contrastive loss is optimized for both real and TTS features, with  additional auxiliary losses on TTS material. %

\subsection{Training on Unspoken Text and Untranscribed Speech}
\label{sec:others}

A major challenge in using TTS for ASR data augmentation is encouraging effective generalization from synthetic to real speech~\cite{wang2020improving,rosenberg2019speech,li2018training}.
Synthesized speech exhibits much less variation than real speech; it has a low SNR, and contains no disfluencies and few internal silences. 
The following two design choices encourage effective self-supervised pretraining from synthetic speech.
First, we mix synthetic and real utterances within each batch.  A loss mask $\sigma$ is then used to combine the speech and text based losses as follows
\begin{eqnarray}
\mathcal{J} = \sigma\cdot \mathcal{J}_{\tt text} + (1-\sigma)\cdot\mathcal{J}_{\tt speech}
\end{eqnarray} This forces the model to learn representations that are effective for both synthetic and real speech.
Second, we apply data augmentation the synthetic speech when optimizing the auxiliary losses.
The $\mathcal{J}_{\tt w2v} (x\mid \theta_e)$ necessarily includes time masking in its loss calculation \cite{baevski2020wav2vec,zhang2020pushing}.   However, for $\mathcal{J}_{\tt aux}$ on TTS data, we use SpecAugment \cite{park2019specaugment}, applying {\bf both} time and frequency masking. 
SpecAugment frequency masking promotes better generalization from synthetic to real speech (cf.~Section \ref{sec:aug}).

\section{Data}
\label{sec:data}

\begin{table*}[t]
  \caption{Description of Supervised (Sup.)/Unsupervised (Unsup.) speech and text in various corpora}
  \vskip 0.1in
  \label{tab:data}
  \centering
  \resizebox{0.95\width}{!}{%
  \begin{tabular}{lcccc}
    \toprule
    Corpus & Sup. Audio data (hrs) & Unsup. Audio (hrs) & Unsup. Text (utts) & Text data selection \\
    \midrule
    Librispeech     & 960 & Librilight-60k & 40M  & No\\
    AMI             & 100 & Librilight-60k & 43.5M  & No\\
    \midrule
    In-house en-us & 1000 & Librilight-60k or Youtube-1M & 100M  & Yes\\
    In-house mr-in & 16k  & Youtube-1.7M & 20M  & Yes\\
    \bottomrule
  \end{tabular}
  }
\end{table*}

\myparagraph{ASR:}
The training and test data sets used in this paper including public well-benchmarked corpora and in-house voice search corpora. These are detailed in Table~\ref{tab:data}. The first two rows  correspond to the three public corpora, LibriSpeech~\cite{panayotov2015librispeech}, LibriLight~\cite{kahn2020libri} and AMI~\cite{carletta2005ami}. The last two rows describe two in-house data sets representative of Google's voice search (VS) traffic in two languages, U.S. English (en-us) and Marathi (mr-in). The in-house ASR training data from voice search utterances for both languages are anonymized and hand-transcribed. The development and test sets are a small fraction of training set held out for validation and evaluation. The unspoken text used in pretraining, labeled as {\it Unsup. Text} in  Table~\ref{tab:data} comprises of anonymized and aggregated, typed search query data. These text queries were selected from a much larger pool of 2000M and 170M queries for English and Marathi respectively using the data selection method described in Section~\ref{sec:tts}. In addition, to measure the long-tail word performance in voice search queries, a 15k synthetic test set~\cite{peyser2020improving} targeting rare proper nouns or words with surprising  pronunciations is used. It is important to note that the TTS model used to generate this synthetic test set not only uses a different architecture from what is used in pretraining, but also has no speaker overlap.

\begin{table}[htbp]
  \caption{ASR model parameters: Encoder has 1024 dim. and decoder is a 2-layer LSTM with 1024 cells.}
  \vskip 0.1in
  \label{tab:model_params}
  \centering
  \small
  \resizebox{0.9\width}{!}{%
  \begin{tabular}{lcccccccc}
    \toprule
    Corpus & \bfseries Model & \# Params~(B) & \makecell{\# Conformer\\Layers} & \makecell{Relative\\Attention} \\
    \midrule
    Librispeech/AMI & XL     & 0.6 & 24  & N \\
    Librispeech     & XXL    & 1.0 & 42   & N \\
    \midrule
    In-house        & XL   & 0.6 & 2  & Y \\
    \bottomrule
  \end{tabular}
  }
  \vspace{-1em}
\end{table}

\myparagraph{TTS:}
Two different TTS corpora are used in this paper. We use the freely available LibriTTS~~\cite{zen2019libritts} corpus containing a total of 960 hours of segmented Librispeech data from 2,456 speakers. %
The second corpus is an in-house 30-hour data set comprising of 7 Marathi professional speakers.

\myparagraph{Language Model:}
The Librispeech text corpus comprises of nearly 803 million tokens from 40M utterances of filtered text derived from 14.5K Project Gutenberg books~\cite{panayotov2015librispeech}. Training data for the Voice Search experiments are randomly drawn from a number of text sources including supervised transcripts used in E2E model training, YouTube search logs, Google search queries, Maps search queries and crawled web documents~\cite{biadsy2017effectively,tara2021}. Overall, this amounts to nearly 100 billion and 380 million text sentences for English and Marathi respectively.

\begin{table*}[htbp]
  \caption{WERs(\%) when using the LibriSpeech~960hr as supervised data. We compare models trained without any unlabeled data~(Row 1) and fine-tuned from a pretrained model using supervised data~(Pretraining). We include the best results of several methods in the literature, and their corresponding references are where the numbers are quoted from. The lowest WER(s) under different settings are marked in bold.}
  \vskip 0.1in
  \label{tab:960hr}
  \centering
  \resizebox{0.9\width}{!}{%
  \begin{tabular}{lcccccccccc}
    \toprule
    \bfseries Method & \multirowcell{2}{\\[-7pt]Unlabeled\\Data~(hrs)}
    & \multirowcell{2}{\\[-7pt]AM\\Size~(B)} & \multicolumn{4}{c}{\bfseries No LM} & \multicolumn{4}{c}{\bfseries With LM} \\
    \cmidrule(r){4-7} \cmidrule(r){8-11}
    & & & \bfseries dev & \bfseries dev-other & \bfseries test & \bfseries test-other
     & \bfseries dev & \bfseries dev-other & \bfseries test & \bfseries test-other \\
    \midrule
    \bfseries Random Initialization \\
    \quad  Conformer~L~\cite{zhang2020pushing}
    & N/A & 0.1 
    & 1.9 & 4.4 & 2.1 & 4.3
    & $-$ & $-$ & 1.9 & 3.9 \\
    \midrule
    \bfseries Pretraining audio only\\
    \quad wav2vec~2.0~\cite{baevski2020wav2vec}
    & 60k & 0.3 
    & 2.1 & 4.5 & 2.2 & 4.5
    & 1.6 & 3.0 & 1.8 & 3.3 \\ 
    \quad HuBERT~Large~\cite{hsu2021hubert}
    & 60k & 0.3 
    & $-$ & $-$ & $-$ & $-$
    & 1.5 & 3.0 & 1.9 & 3.3 \\
    \quad HuBERT~X-Large~\cite{hsu2021hubert}
    & 60k & 1.0 
    &  $-$ & $-$ & $-$ & $-$
    &  1.5 & \textbf{2.5} & 1.8 & \textbf{2.9} \\
    \quad w2v-Conformer~XL~\cite{zhang2020pushing}
    & 60k & 0.6 
    & 1.7 & 3.5 & 1.7 & 3.5
    & 1.6 & 3.2 & \bfseries 1.5 & 3.2 \\
    \quad w2v-Conformer~XXL~\cite{zhang2020pushing}
    & 60k & 1.0 
    & 1.6 & 3.2 & 1.6 & 3.3 
    & 1.5 & 3.0 & \bfseries 1.5 & 3.1 \\
    \bfseries Pretraining audio and text\\
    \quad \cite{zhang2020pushing}+tts4pretrain~XL~(Ours)
    & 60k & 0.6 
    &  1.6 & 3.4 &  1.6 & 3.2
    & \bfseries 1.5 & 3.1 &  \textbf{1.6} & 3.1 \\
    \quad \cite{zhang2020pushing} +tts4pretrain~XXL~(Ours) 
    & 60k & 1.0 
    & \textbf{1.5} & \textbf{3.0} & \textbf{1.6} & \textbf{3.0}
    & \textbf{1.5} & 2.8 & \textbf{1.5} & \textbf{2.9} \\
    \bottomrule
  \end{tabular}
  }
  \vspace{-1em}
\end{table*}

\section{Model Descriptions}
\label{sec:setup}
\subsection{ASR}
The ASR network is a RNN transducer \cite{graves2012sequence} consisting of a LSTM decoder and a Conformer encoder~\cite{gulati2020conformer}. The encoder is a stack of "conformer block"s, each of which is a series of multi-headed self attention \cite{vaswani2017attention}, depth-wise convolution and feed-forward layers. The model configuration  is summarized in Table~\ref{tab:model_params}. All models are trained on use 80-dimensional log-mel filter bank coefficients. The experiments on the public corpora use 1024 word-piece targets and the Voice Search experiments use 4K word-piece targets~\cite{kudo2018sentencepiece}.

\textbf{Pretraining Parameters:} %
All ASR models are trained on Google TPU V3 cores~\cite{jouppi2020domain}. For the XL model, we use Adam optimization and cap the norm of the gradient to 20. For the XXL model, we switch the optimizer to Adafactor~\cite{adafactor} with $\beta_1=0.9$ and $\beta_2=0.98$, and use 2nd-moment estimator factorization to reduce the accelerator's memory footprint. Both models use a transformer learning rate schedule in~\cite{vaswani2017attention} with a peak learning rate of 2e-3 and 25k warm-up steps. For {\it tts4pretrain}, we use the same settings, with a global batch size of 1024 and 5 times less learning rate. The global batch size used for corpora $\geq 1000h$ is 512 while for corpora $\leq 1000h$ is 256. We introduce phoneme and word-piece auxiliary decoders described in Section~\ref{sec:proposed} and explore CTC and RNNT as two options for the training objective. The CTC objective uses a single layer prediction network while the RNN-T objective uses a 2-layer LSTM network.

\textbf{Fine-tuning Parameters:} Following \cite{zhang2020pushing} we optimize the encoder and decoder with separate optimizers and learning rate schedules. We use Adam optimization with the transformer learning rate schedule described in~\cite{zhang2020pushing}. The encoder uses a peak learning rate of 3e-4 with 5k warm-up steps. The decoder uses a peak learning rate of 1e-3 and 1.5k warm-up steps. For evaluation, we keep a separate copy of exponential-moving-averaged model weights aggregated with decay rate 0.9999.  %

\subsection{TTS}
The multi-speaker TTS model uses a Tacotron2 TTS architecture described in \cite{wang2020improving} with hierarchical VAE~\cite{hsu2018hierarchical}.
The input sequence embedding is encoded by three
convolutional layers, which contain 512 filters with shape 5 x 1,
followed by a bidirectional long short-term memory (LSTM)
layer of 256 units for each direction. The resulting embeddings are accessed by the decoder through a location sensitive
attention mechanism.
The decoder is followed by a PostNet with five convolutional layers of 512 filters with shape 5 x 1. 

\subsection{Language Model}
The Librispeech LM is an eight-layer 103M-parameter transformer language model~\cite{zhang2020pushing} trained on the LibriSpeech language model corpus~\cite{panayotov2015librispeech}. The in-house en-us Voice Search experiments use a Conformer LM described in~\cite{biadsy2017effectively, tara2021} trained on multiple domains. The in-house mr-in Voice Search experiments use an N-gram LM for 1-pass decoding and a maximum-entropy LM~\cite{biadsy2017effectively} for 2-pass rescoring.  During rescoring, the first-pass LM’s log-likelihood is log-linearly interpolated
with the second-pass model score~\cite{biadsy2017effectively}.

\section{Results}
\label{sec:exp}

\subsection{Librispeech}
Table~\ref{tab:960hr} presents our results on LibriSpeech evaluation sets when using the~960hr supervised training corpus. The TTS model used in the section is trained with the LibriTTS data described in Section~\ref{sec:data}. We compare a number of state-of-the-art self-supervised representation learning methods from the literature including, the recently introduced techniques, HuBERT~\cite{hsu2021hubert} and w2v-Conformer~\cite{zhang2020pushing}.

As shown in Table~\ref{tab:960hr}: 1) Injecting text information does help train a better speech encoder for ASR: Even without an external LM, a model pretrained with {\it tts4pretrain} matches other models. To the best of our knowledge, after LM fusion, it has resulted in a new state-of-the-art baseline with the Librispeech 960 hour with pretrain only model (last row);
2) Larger model benefits more from text data: The relative gain from 1B model is larger than 600M model. We believe a larger model has increased capacity to better utilize the text information;
3) Representations learned from speech and text during pretraining is better than speech alone: As shown in the table, {\it tts4pretrain} based pretraining reaches the same performance as a speech-only pretrained model coupled with external LM fusion. We hypothesize that the encoder has now learned contextual, lexical information. In order to get additional wins from LM fusion, the LM would have to be trained on either different text sources or spanning  domains not included in pretraining.

\subsection{AMI Meeting Transcription}
\begin{table}[thbp]
  \vspace{-2em}
  \caption{\label{tab:exp-ami} {Performance of {\it tts4pretrain} on AMI individual headset microphone (ihm) and single distant microphone (sdm1) test sets.} }
  \vspace{1em}
  \centerline{
    \begin{tabular}{m{14em}   c c }
      \toprule
     Method & {\em ihm} &{\em sdm} \\
      \midrule
      Baseline conformer XL &26.1 &40.5 \\
      \midrule
\ \  + Libri-Light pretrain &	10.7& 24.7\\
\ \ \ \  + {\bf \footnotesize{tts4pretrain SpeechStew Text}} & 10.1&	24.0\\
      \midrule
\ \ \ \ + \small Supervised SpeechStew~\cite{chan2021speechstew} & 9.6 & 23.8	\\
\bottomrule
    \end{tabular}
  }
  \vspace{-2em}
\end{table}
Table~\ref{tab:exp-ami} presents results on the AMI meeting transcription task.  Speech-only pretraining on Libri-Light followed by fine-tuning on the AMI corpus provides significant reduction in WER (Row 2). {\it tts4pretrain} provides an additional 5.6\% relative win over the speech-only pretrained model (Row 3). We present an additional data point for {\it tts4pretrain} by incorporating text from the supervised transcripts in other freely-available corpora as the {\it unspoken text} corpus. This yields an additional 3.5M utterances (over the 40M from Librispeech) from the SpeechStew~\cite{chan2021speechstew} training set, a combination of 7 publicly available supervised speech corpora. The last row in Table~\ref{tab:exp-ami} serves as a reference baseline when a model is trained with all the available corpora as done in~\cite{chan2021speechstew}. From Row 3, we observe that textual information in pretraining can compensate for lack of real acoustic training data as {\it tts4pretrain} is able to close the gap with the reference baseline performance.

\subsection{Voice Search}

We begin with results on English Voice Search queries. The TTS model for {\it tts4pretrain} was trained using the LibriTTS corpus described in Section~\ref{sec:data}. Tables~\ref{tab:exp-enus} and ~\ref{tab:exp-enus-t} present results comparing {\it tts4pretrain} with audio-only self supervision on two different test sets described in Section~\ref{sec:data}. The first row in Table~\ref{tab:exp-enus} illustrates the performance of the baseline model with and without LM fusion. It can be seen that \textit{tts4pretrain} improves over audio-only pretraining by 15\% relative (Row 3). When trained with 15-fold more YouTube data (last row), \textit{tts4pretrain} still improves over audio-only pretraining by 10\% relative.  It is interesting to note from Rows 3 and 5 in Table~\ref{tab:exp-enus}, that with less untranscribed speech, LM fusion seems less effective after \textit{tts4pretrain}. Table~\ref{tab:exp-enus-t} shows a similar trend on two rare word testsets . The integration of an external LM is more crucial for recognizing rare words than commonly used words and is  consistent with the observations in~\cite{peyser2020improving}.

\begin{table}[thbp]
  \vspace{-1em}
  \caption{\label{tab:exp-enus} {Results on en-us Voice Search queries with {\it tts4pretrain} and LM fusion} }
  \vspace{1em}
  \centerline{
    \begin{tabular}{m{10em}   c c  }
      \toprule
     \multirow{2}{8em}{{Pretraining Data}} &  \multicolumn{2}{c}{Search WER(\%)}\\
     & w/o LM & w/ LM  \\
      \midrule
None&10.7& 9.0\\
\midrule
Libri-Light speech &7.3&6.5 \\
\ \  + {\bf tts4pretrain}&6.2&6.1 \\
\midrule
Youtube speech (15X)&6.7&6.1 \\
\ \  + {\bf tts4pretrain}&6.1&5.6 \\
\bottomrule
    \end{tabular}
  }
  \vspace{-2em}
\end{table}

\begin{table}[thbp]
  \vspace{-1em}
  \caption{\label{tab:exp-enus-t} {Rare word performance} }
  \vspace{1em}
  \centerline{
    \begin{tabular}{m{8em}   c c  }
      \toprule
     \multirow{2}{8em}{{Pretraining Data}} &  \multicolumn{2}{c}{Rare WER(\%)}\\
     & w/o LM & w/ LM  \\
      \midrule
None& 38.8 & 32.5\\
\midrule
Libri-Light speech&34.2 &27.4 \\
\ \  + {\bf tts4pretrain}&31.2& 25.2 \\
\bottomrule
    \end{tabular}
  }
  \vspace{-1em}
\end{table}

\begin{table}[htbp!]
  \vspace{-1em}
  \caption{\label{tab:exp-mr} {Results on Marathi Voice Search queries with {\it tts4pretrain} and LM fusion} }
  \vspace{1em}
  \centerline{
    \begin{tabular}{m{10em}   c c  c }
      \toprule
    Method  &RNNT &HAT &HAT $\circ$ LM\\
      \midrule
Baseline conformer XL&20.7&20.1&19.2\\
\ \ \ \ + Speech pretrain&20.3&19.8&18.8\\
\ \ \ \ \ \ \ \ + {\bf tts4pretrain}&19.6&19.3&18.6\\
\bottomrule
    \end{tabular}
  }
  \vspace{-1em}
\end{table}

Next, we present results on Marathi voice search queries in Table~\ref{tab:exp-mr}. The TTS model for {\it tts4preTrain} was trained using the Marathi TTS corpus described in Section~\ref{sec:data}. We study LM integration with a Hybrid Autoregressive Transducer (HAT) model. The HAT model couples the powerful modeling ability of E2E models with an inference algorithm that separately preserves the internal LM learned by the E2E model thus allowing for integration with an external LM. We observe that {\it tts4pretrain} outperforms audio-only self-supervision by 4\% relative, with the best performing model being the HAT model at 18.6\% WER. We observe that the gains from all the methods presented in Table~\ref{tab:exp-mr} are a lot less than those seen in English queries. We attribute this to the increased amount of real-speech used in Marathi for pretraining and fine-tuning but 5-fold less unspoken text compared to English queries. The performance of {\it tts4pretrain} in these two languages provide insight into the impact of pretraining with varied amounts of unspoken text and untranscribed speech.

\section{Analysis}
\label{sec:analysis}
In this Section, we explore several questions to better understand the impact and  behavior of {\it tts4pretrain}.

\subsection{How much unsupervised data do we need?}
In order to answer the question on the amount of unsupervised data needed to leverage {\it tts4pretrain}, we look at both untranscribed speech and text.
All experiments in this section are conducted on English Voice Search queries.

Table~\ref{tab:speech-amount} includes the performance of an ASR model pretrained with different amounts of untranscribed speech and a fixed amount 100M of unspoken text.  The first row shows how the model improves with increasing amounts of speech by speech-only pretraining using wav2vec2.0. While there is a significant improvement in performance with every 10-fold increase in data, it can be seen that the gains begin to asymptote, with a 26\% relative win from 600-hour to 6000-hours of pretraining, followed by only 9.8\% relative gains when increasing the training data to 60K hours. The second row presents the same analysis with {\it tts4pretrain}. Here, we see a more uniform trend with approximately 10\% relative win in both 10-fold increases of data. It can also be seen that there is no WER reduction (a small regression exists) seen with wav2vec2.0 when training with  600 hours of real speech. However, when the same amount of speech is supplemented with synthesized speech by {\it tts4pretrain}, a 26\% relative gain can be seen with 10-fold less speech data. This suggests that the combination of speech and text modalities is effective and particularly useful for languages where less real speech is available. With subsequent additions of real speech, the model is able to learn more effectively and outperform speech-only pretraining.

\begin{table}[thbp!]
\vspace{-2mm}
  \caption{\label{tab:speech-amount} {The amount of untranscribed speech in pretraining.} }
  \vspace{1em}
  \centerline{
    \begin{tabular}{m{7em}   c c  c c }
      \toprule
Pretrain data&\multicolumn{4}{c}{Untranscribed speech amount}\\
&0 hr&600 hr&6k hr&60k hr\\
      \midrule
Speech &10.7&11&8.1&7.3\\
\ \ \ \ + Text&-&7.8&7&6.2\\
  \bottomrule
    \end{tabular}
  }
\end{table}

Next, we explored the impact of the amount of unspoken text injected via TTS while keeping the amount of untranscribed speech at 60K hours. Table~\ref{tab:text-amount} shows that the initial addition of 1M utterances yields a win of 6.8\% relatve. However, the next similar win requires a 100-fold increase in the amount of unspoken text. While not a surprising result, it offers insight in balancing the needs and costs of acquiring untranscribed speech and unspoken text.

\begin{table}[thbp!]
\vspace{-2mm}
  \caption{\label{tab:text-amount} {The amount of unspoken text in pretraining.} }
  \vspace{1em}
  \centerline{
    \begin{tabular}{m{7em}   c c  c c }
      \toprule
Unspoken text & None&1M&10M&100M\\
      \midrule
WER & 7.3&6.8&6.6&6.2\\
  \bottomrule
    \end{tabular}
  }
\end{table}

We observed from Table~\ref{tab:exp-enus} that regardless of the style of speech, Librispeech or YouTube videos, the WER on this task reached the same 6.2\% wth {\it tts4pretrain}. Table~\ref{tab:text-domain} studies the effect of domain mismatch in unspoken text.  The use of Librispeech LM text in pretraining does not provide as much gain (7.0\%) as typed text queries (6.5\%) which are better matched to the Voice Search task. An additional modest win can be obtained (6.2\%) with data selection described in Section~\ref{sec:tts} to better match the domain to the task at hand.

\begin{table}[thbp!]
\vspace{-2mm}
  \caption{\label{tab:text-domain} {Unspoken text selection for pretraining.} }
  \vspace{1em}
  \centerline{
    \begin{tabular}{m{14em}   c  }
      \toprule
Pretraining Data&Search WER\\
      \midrule
Librilight&7.3\\
\midrule
\ \ \ \ + Librispeech LM training text&7.0\\
\ \ \ \ + random typed queries &6.5\\
\ \ \ \ \ \ \ \ + text data selection&6.2\\
\bottomrule

    \end{tabular}
  }
  \vspace{-2mm}
\end{table}

\subsection{Impact of training objective and auxiliary decoders}
In this Section, we explore few obvious choices for the training objective and  auxiliary decoders. These ablation studies were conducted on the smaller 100-hour supervised Librispeech corpus and 60K hours of unsupervised pretraining. As mentioned in Section~\ref{sec:proposed}, we explored two different training objectives for the decoder  in {\it tts4pretrain}.  Table~\ref{tab:ctc-rnnt} concludes that a CTC loss based decoder works better than an RNN-T decoder. We attribute this to the better alignment properties of CTC compared to RNN-T.
 Table~\ref{tab:decoder-loss} shows that without any type of auxiliary decoder to enforce lexical information, the model is able to learn very little from the synthesized speech alone. All experiments in this table used a CTC training objective based on the conclusion from Table~\ref{tab:ctc-rnnt}.  Introducing auxiliary decoders with word-piece and phonemic targets (which come for free from the TTS front-end) improves learning from unspoken text with the best result (last row) obtained by using both objectives.

\label{sec:decoders}
\begin{table}[thbp!]
\vspace{-2mm}
  \caption{\label{tab:ctc-rnnt} {Decooder training objective used in {\it tts4pretrain}} }
  \vspace{1em}
  \centerline{
    \begin{tabular}{m{10.5em}   c c  c c }
      \toprule
 \em{LS 100h System}&dev&devother&test&testother \\
      \midrule
Libri-Light + tts4retrain&2.5&4.9&2.5&4.9\\
\ \ \ \ + RNNT loss&2.4&4.7&2.4&4.7\\
\ \ \ \ + {\bf CTC loss}&2.3&4.7&2.3&4.7\\
  \bottomrule
    \end{tabular}
  }
  \vspace{-1em}
\end{table}

\begin{table}[thbp!]
  \caption{\label{tab:decoder-loss} {Impact of additional auxiliary decoders in {\it tts4pretrain} } }
  \vspace{1em}
  \centerline{
    \begin{tabular}{m{10em}   c c  c c }
      \toprule
 \em{LS 100h System}&dev&devother&test&testother \\
      \midrule
Libri-Light pretrain  &2.6&4.8&2.6&5\\
\midrule
\ \ + tts4pretrain &2.5&4.9&2.5&4.9\\
\ \ \ \ + wpm loss&2.3&4.7&2.3&4.7\\
\ \ \ \ \ \ + {\bf phoneme loss}&2.2&4.5&2.3&4.6\\
\bottomrule
    \end{tabular}
  }
  \vspace{-1em}
\end{table}

\subsection{Impact of data augmentation on the synthesized speech}
\label{sec:aug}

Synthesized speech that has been augmented with different noise styles is effective in robust model training~\cite{chen2020unpair}.  We present different masking schemes used to augment  TTS data during pretraining in Table~\ref{tab:aug}.  We find that the 50\% time masking used in wav2vec2.0 is not optimal for ASR-derived losses on TTS utterances in {\it tts4pretrain}. The best setup uses 20\% time and frequency masking with frequency warping. This is consistent with SpecAugment hyperparameters used in downstream ASR~\cite{park2019specaugment}. Note, this augmentation is only used for the auxiliary, decoder loss not the contrastive loss. 
\begin{table}[thbp!]
  \caption{\label{tab:aug} {Data augmentation on synthesized speech} }
  \vspace{1em}
  \centerline{
    \begin{tabular}{m{10em}   c c  c c }
      \toprule
Augmentation Type &dev&devother&test&testother\\
      \midrule
50\% time mask &2.5&5.2&2.6&4.9\\
20\% time+freq. mask&2.5&5.2&2.4&4.9\\
\ \ \ \ + {\bf freq. warp}& 2.4&5.1&2.3&4.8\\
  \bottomrule
    \end{tabular}
  }
  \vspace{-1em}
\end{table}

\section{Conclusion}
\label{sec:conclude}
We propose {\it tts4pretrain}, a method to learn self-supervised representations from both untranscribed speech and unspoken text using 1) speech synthesis to generate speech from unspoken text and 2) auxiliary decoders and losses based on ASR objectives for this synthesized speech.  
\textit{tts4pretrain} yields WER reductions of 10\% relative on the well-benchmarked, Librispeech task over a state-of-the-art baseline pretrained with wav2vec2.0 only.  The effectiveness of \textit{tts4pretrain} is also demonstrated on AMI and in-house data. We show that \textit{tts4pretrain} is effective on different encoder architectures and sequence training objectives such as CTC, RNN-T, and HAT.  Moreover, language-model fusion is shown to be complementary to the introduction of textual information via \textit{tts4pretrain}.

\bibliographystyle{IEEEbib}
\bibliography{strings,refs}

\end{document}